\documentclass[11pt]{article}

\usepackage[preprint]{acl}
\usepackage{times}
\usepackage{latexsym}
\usepackage{amsfonts}
\usepackage{amsmath}
\usepackage{amssymb}
\usepackage{multicol}
\usepackage{multirow}
\usepackage{xspace}
\usepackage{booktabs}
\usepackage{bbding}
\usepackage{array}
\usepackage{threeparttable}
\usepackage{tabularx}
\usepackage{enumitem}
\usepackage[linesnumbered,ruled,vlined]{algorithm2e}
\usepackage{colortbl}
\usepackage{color}
\usepackage{setspace}
\usepackage{makecell}
\usepackage{listings}
\usepackage{xltabular}
\usepackage{tcolorbox}

\usepackage{xcolor}
\definecolor{JungleGreen}{RGB}{41,171,135}
\definecolor{Cerulean}{RGB}{0,123,167}
\definecolor{CornflowerBlue}{RGB}{100,149,237}
\definecolor{Black}{RGB}{0,0,0}
\tcbuselibrary{breakable}
\usepackage{cleveref}
\crefformat{section}{\S#2#1#3}
\crefformat{subsection}{\S#2#1#3}
\crefformat{subsubsection}{\S#2#1#3}
\crefrangeformat{section}{\S\S#3#1#4 to~#5#2#6}
\crefmultiformat{section}{\S\S#2#1#3}{ and~#2#1#3}{, #2#1#3}{ and~#2#1#3}
\crefmultiformat{subsection}{\S\S#2#1#3}{ and~#2#1#3}{, #2#1#3}{ and~#2#1#3}

\Crefformat{figure}{#2Fig.~#1#3}
\Crefmultiformat{figure}{Figs.~#2#1#3}{ and~#2#1#3}{, #2#1#3}{ and~#2#1#3}
\Crefformat{table}{#2Tab.~#1#3}
\Crefmultiformat{table}{Tabs.~#2#1#3}{ and~#2#1#3}{, #2#1#3}{ and~#2#1#3}
\Crefformat{appendix}{Appx.~\S#2#1#3}
\crefmultiformat{appendix}{Appx.~\S#2#1#3}{ and~#2#1#3}{, #2#1#3}{ and~#2#1#3}
\crefformat{algorithm}{Alg.~#2#1#3}
\Crefformat{equation}{Eq.~#2#1#3}
\newcommand{\xhdr}[1]{\vspace{0.3em}\noindent{{\bf #1.}}}

\newenvironment{colblock}[1][\textbf{Collary}]{\begin{tcolorbox}[colback=JungleGreen!10!Cerulean!10,colframe=CornflowerBlue!60!Black,title = \textbf{#1}, breakable, boxrule=1pt]}{\end{tcolorbox}}

\usepackage[T1]{fontenc}

\usepackage[utf8]{inputenc}

\usepackage{microtype}

\usepackage{inconsolata}

\title{Language Models can Evaluate Themselves via Probability Discrepancy}

\author{Tingyu Xia$^{1,3}$\footnotemark[2] \quad  Bowen Yu$^{2}$\footnotemark[1] \quad Yuan Wu$^{1,3}$\footnotemark[1] \quad Yi Chang$^{1,3,4}$\footnotemark[1] \quad Chang Zhou$^{2}$\\
        $^{1}$School of Artificial Intelligence, Jilin University \\
        $^{2}$Alibaba Group \\
        $^{3}$Engineering Research Center of Knowledge-Driven Human-Machine Intelligence, MOE, China \\
        $^{4}$International Center of Future Science, Jilin University\\
        xiaty21@mails.jlu.edu.cn, yubowen.ybw@alibaba-inc.com \\
	 yuanwu@jlu.edu.cn, yichang@jlu.edu.cn, ericzhou.zc@alibaba-inc.com \\       
}
\begin{document}
\maketitle
\renewcommand{\thefootnote}{\fnsymbol{footnote}}
\footnotetext[2]{Work done during the author's internship at the Alibaba Group}
\footnotetext[1]{Corresponding authors}
\renewcommand{\thefootnote}{\arabic{footnote}}

\begin{abstract}

In this paper, we initiate our discussion by demonstrating how Large Language Models (LLMs), when tasked with responding to queries, display a more even probability distribution in their answers if they are more adept, as opposed to their less skilled counterparts. Expanding on this foundational insight, we propose a new self-evaluation method ProbDiff for assessing the efficacy of various LLMs. This approach obviates the necessity for an additional evaluation model or the dependence on external, proprietary models like GPT-4 for judgment. It uniquely utilizes the LLMs being tested to compute the probability discrepancy between the initial response and its revised versions. A higher discrepancy for a given query between two LLMs indicates a relatively weaker capability. Our findings reveal that ProbDiff achieves results on par with those obtained from evaluations based on GPT-4, spanning a range of scenarios that include natural language generation (NLG) tasks such as translation, summarization, and our proposed Xiaohongshu blog writing task, and benchmarks for LLM evaluation like AlignBench, MT-Bench, and AlpacaEval, across LLMs of varying magnitudes.  The code is available at \url{https://github.com/xiatingyu/ProbDiff}.

\end{abstract}
\section{Introduction}

With the emergence of LLMs like ChatGPT, we've witnessed groundbreaking progress in tasks involving instruction following~\cite{wang2023aligning}, intent comprehension~\cite{lu2023instag}, and text generation~\cite{zhao2023survey}. As LLMs evolve at a rapid pace, it becomes crucial to develop solid evaluation frameworks to gauge their performance accurately~\cite{chang2023survey}. Traditional evaluation methods such as BLEU~\cite{papineni2002bleu} focus mainly on superficial text differences and often fail to align with human judgment~\cite{liu2023gpteval}.

\begin{figure}[htbp]
\centering
\includegraphics[width=\columnwidth]{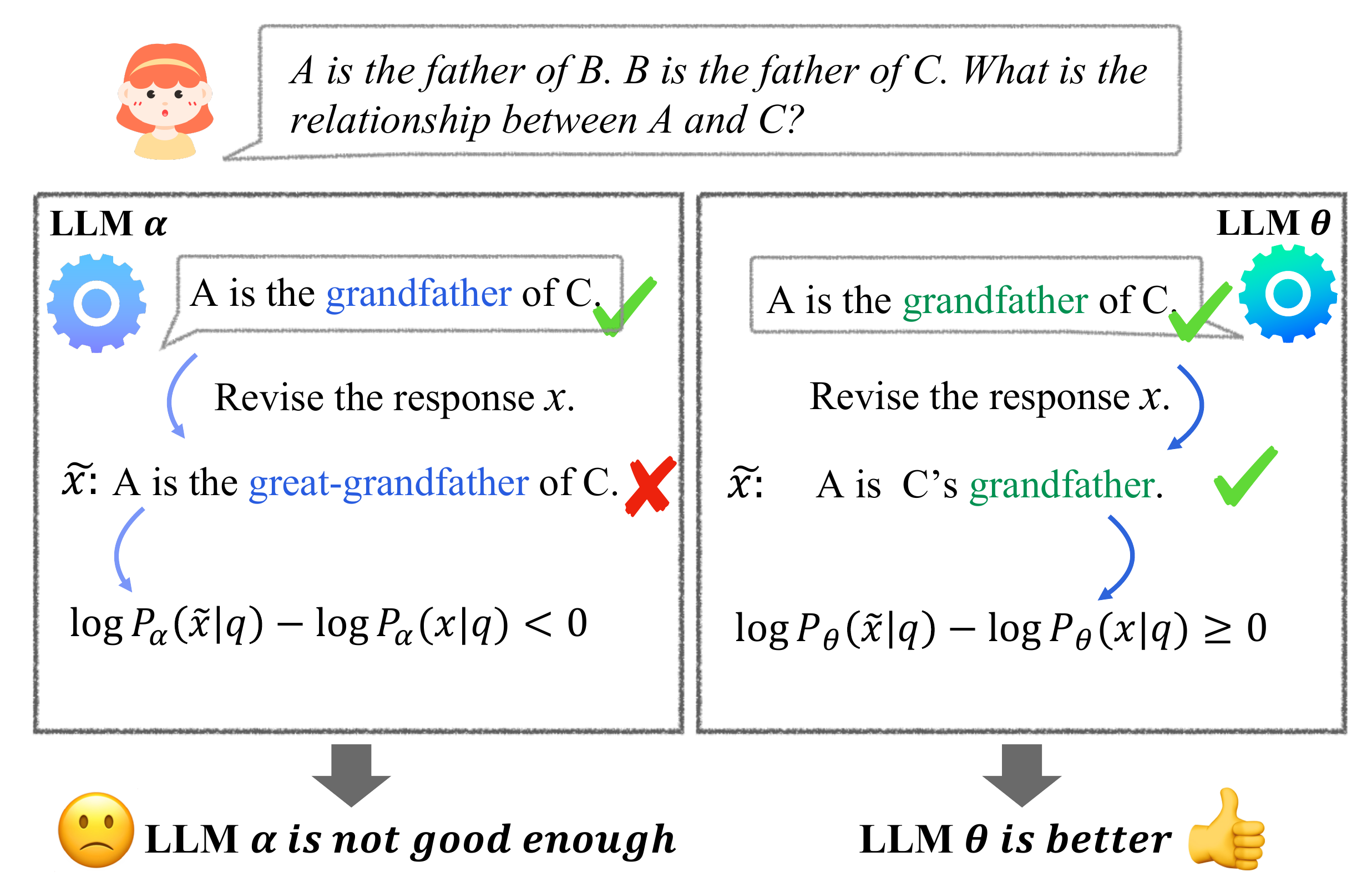}
\caption{An overview of \textbf{ProbDiff}, wherein an LLM iteratively revises its responses, calculating the resulting probability discrepancies as an self-evaluation metric. Larger discrepancies imply decreased confidence in the generated outcomes, with greater variances indicating poorer performance.}
\label{fig:model_struc}
\vspace{-0.5em}
\end{figure}

Critique models, trained on binary preference and language
critique datasets to simulate human judgment~\cite{ouyang2022training}, have emerged as a core component of Learning from Human Feedback~\cite{stiennon2020learning}. However, critique models encounter several challenges. Firstly, they risk becoming outdated as LLMs must continually adapt to novel and intricate tasks, such as writing in specific styles like Xiaohongshu, or solving increasingly complex problems like tool use~\cite{li2023api} and role play~\cite{lu2024large}. 
Frozen critique models trained on past data struggle to accurately assess such unseen tasks, and it's impractical to predefine and annotate data for all potential tasks in labeled feedback data. Secondly, there is the issue of \textit{Reward Hacking}, where an LLM may exploit vulnerabilities in the critique models, achieving high rewards without truly fulfilling the intended objectives~\cite{askell2021general,rame2024warm}, resulting in degraded performance, such as linguistically flawed~\cite{lewis2017deal} or excessively verbose outputs~\cite{singhal2023long}. Thirdly, with each iteration and upgrade of an LLM, there is a need to train additional critique models of matching or greater capacity, increasing the costs and resource consumption.
Another popular option is to allow proprietary models like GPT-4 to score candidate responses via API~\cite{chen2023exploring, fu2023gptscore,ji2023exploring,kocmi2023large,wang2023chatgpt,zhang2023wider}. However, GPT-based evaluations entail the risk of potential high costs associated with addressing data leaks~\cite{wang2023pandalm}, and are subject to OpenAI's terms of use~\footnote{\url{https://openai.com/policies/terms-of-use}}.
Overall, there currently exists no evaluation method that can fit human preferences without increasing additional training costs, maintain robustness without leaking data, and continually improve alongside advancements in model capabilities.

This paper poses a novel hypothesis: for any query $q$, an LLM capable of providing accurate response $x$ tend to exhibit a more uniform probability distribution ${\rm log}\; p(x|q)$ compared to their less proficient counterparts. Existing studies indicate that the initial model response $x$ often converges to local maxima of the log probability. 
when refining $x$ to generate a new response $\hat{x}$, the relative reduction in ${\rm log}\; p(\hat{x}|q)$ compared to ${\rm log}\; p(x|q)$ is insignificant. 
Conversely, for LLMs incapable of producing a correct response, ${\rm log}\; p(x|q)$ typically demonstrates higher variance. As a result, revised answers tend to exhibit much lower log probabilities than the original response. 
We empirically validate this hypothesis and observe its consistency across various LLMs and tasks. 
Figure~\ref{fig:log_prob_on_alpacaeval} and~\ref{fig:log_curve} illustrate the visualization of the underlying hypothesis and Section~\ref{sec:hypothesis_validation} showcases the empirical assessment.
Exploiting this insight, we introduce ProbDiff, a self-evaluation technique applicable to any LLM across tasks.
Given a query $q$, ProbDiff first prompts a candidate LLM to generate a response $x$, then asks the LLM to revise $x$ based on $q$, producing a refined response $\hat{x}$. Finally, ProbDiff quantifies the probability discrepancy ${\rm log}\;p({\hat x}|q)$ - ${\rm log}\; p(x|q)$  as the evaluation metric. 
When comparing two candidate LLMs on $q$, a larger probability discrepancy indicates a lower proficiency in handling the instruction.

Compared with critique models, ProbDiff utilizes the probability characteristics of LLM to evaluate the LLM's performance, avoiding the resource waste of training a second model and addressing the issues of critique models' inability to dynamically adjust with the improvement of LLM capabilities and task changes. 
Unlike GPT-based Evaluation, ProbDiff mitigates the risks of data leakage, high API call costs, and usage restrictions of external models. 
We conduct experiments on conventional generation tasks such as translation and summarization, novel generation tasks like Xiaohongshu blog writing, as well as English and Chinese LLM benchmarks such as AlpacaEval~\cite{alpaca_eval}, MT-bench~\cite{zheng2023judging}, and AlignBench~\cite{liu2023alignbench}, using Qwen~\cite{bai2023qwen}, LLAMA-2~\cite{touvron2023llama}, Yi~\cite{01yi}, WizardLM~\cite{xu2023wizardlm} and Tulu~\cite{ivison2023camels}. 
The results demonstrate that ProbDiff exhibits highly consistent performance across various tasks and LLMs compared to the currently prevalent GPT-4 based Evaluation methods.

\section{Related Work}
\xhdr{LLM Evaluation}
Current research primarily focuses on model-based evaluation, which involves training another reward model or utilizing external proprietary LLMs such as GPT-4 as judges. Representative methods  of the former like PandaLM~\cite{wang2023pandalm} , Shepherd~\cite{wang2023shepherd}, and AUTO-J~\cite{li2023generative} distinguish the most superior model among various candidates and can provide a brief explanation or critique to support their evaluation.
However, these models are often relatively small in scale, frozen once trained, and unable to adapt to changes in LLM capabilities, output distributions, and task variations.
A more popular approach currently leverages larger and more powerful LLMs like GPT-4 to evaluate numerous natural language generation tasks, including text summarization, machine translation, and so on, showcasing remarkable performance~\cite{liu2023gpteval, wang2023chatgpt, kocmi2023large}. However, subsequent investigations have unveiled certain issues with LLM evaluators, particularly concerning biases related to position and verbosity~\cite{wang2023large,zheng2023judging}. 
Despite researchers adopting multi-agent strategies and making multiple calls to GPT models to mitigate bias~\cite{zhang2023wider,chan2023chateval,li2023prd}, they still face challenges such as data leakage and high evaluation costs.
In this paper, we propose for the first time the concept of probability discrepancy to enable LLMs to conduct self-evaluation from a probabilistic perspective, without requiring training extra models or invoking external ones. 
It allows for the evaluation of model effectiveness on any task using any LLM and achieves results similar to current mainstream GPT-4 evaluations.

\xhdr{Probability Exploitation}
Probability has played a pivotal role in both detecting inconsistencies \cite{she2023cop,jia2023zero} and identifying machine-generated text \cite{tang2023science,mitchell2023detectgpt,bao2023fast}. Zero-shot faithfulness evaluation with a foundation
language model(FFLM), introduced by \citet{jia2023zero}, combines changes in probability based on the premise that pre-adding a text segment consistent with the output increases the likelihood of predicting the desired output. 
Controlling the preference of the generation model with the
help of the prompt (CoP) \cite{she2023cop}, on the other hand, leverages the disparity in probability between a document and a coherent text prompt during inference to detect factual inconsistencies.
In the realm of detecting machine-generated text, DetectGPT~\cite{mitchell2023detectgpt} posits that unlike human-authored text, model-generated text tends to cluster in local maxima of the log probability. Building upon this insight, Fast-DetectGPT \cite{bao2023fast} utilizes conditional probability curvature to reveal disparities in word choices between LLMs and human authors within a given context.
Inspired by DetectGPT, our work further analyzes the local structure of the generation probability function for LLMs capable and incapable of handling specific queries. 
We observed that proficient LLMs exhibit a flatter distribution, whereas less capable ones exhibit a steep distribution. Consequently, we propose evaluating model performance through repeated sampling and assessing the discrepancy in log probability between samples.
Through this novel probabilistic perspective, we successfully offer a new approach to LLM evaluation.

\section{Method}

In this section, we present our self-evaluation protocol, ProbDiff, designed to self-assess the capabilities of LLMs autonomously through logarithmic probability changes.
We begin with preliminary studies to validate the prior assumptions of ProbDiff, followed by a detailed exposition of its intricacies and mechanisms.

\subsection{Hypothesis Validation}
\label{sec:hypothesis_validation}

The first assumption underlying ProbDiff is that when we question an LLM, the more capable it is of providing an answer, the more confident it should be in its response. 
Therefore, the proportion of consistent or approximate answers in repeated questioning of the same question should be higher.
However, when the model lacks proficiency in a specific domain, it should exhibit considerable uncertainty in its responses. Thus, there would be a significant variance in the answers provided upon repeated questioning.
To test this hypothesis, we conducted experiments using two widely used LLMs: GPT-3.5 and GPT-4. It is well-known that GPT-4 surpasses GPT-3.5 in various dimensions of capability.
If our hypothesis holds true, then for the same set of questions asked multiple times to both GPT-4 and GPT-3.5, the similarity between the answers provided by GPT-4 should be higher.

We employed MT-Bench as our test dataset. Since MT-Bench is a multi-turn dataset, we utilized only the first-turn query, and requested these LLMs to provide answers three times. Specifically, upon receiving the first answer from the LLM, we continued to prompt with "Answer this question again." to collect the model's second answer, and so forth until obtaining the third answer.




\begin{table}[ht]
\renewcommand\arraystretch{1.2} 
\setlength{\tabcolsep}{5.5pt}
\centering
\begin{tabular}{l|cccc}
\toprule
\multicolumn{1}{c|}{\multirow{2}{*}{}} 
&\multicolumn{2}{c}{\textbf{GPT-3.5-turbo}} 
&\multicolumn{2}{c}{\textbf{GPT-4}}\\

\multicolumn{1}{c|}{} &Similarity &EM &Similarity &EM\\

\midrule

$x$ - $x_1$ & 86.88 & 21.25 & 92.25 & 26.25 \\
$x$ - $x_2$ & 90.28 & 18.75 & 90.71 & 25.00 \\

\bottomrule
\end{tabular}
\caption{The preliminary study aims to verify the disparities in similarity between responses to the same query from a strong model (GPT-4) and a weak model (GPT-3.5-turbo). Here, $x$ represents the initial response, $x_1$ denotes the re-generated response, and $x_2$ is the response regenerated based on $x_1$. Similarity refers to the similarity score between two responses, while EM stands for Exact Match, indicating the proportion of responses that are identical in the dataset. This study was conducted using the MT-Bench benchmark.} 
\label{tab:gpt_result}
\end{table}


As depicted in Table~\ref{tab:gpt_result}, the similarity between the first and second responses from GPT-3.5, compared to GPT-4, was lower by 5.37 pts. Additionally, we observed that in both GPT-3.5 and GPT-4, a substantial proportion of responses remained exactly the same across the two rounds, constituting an exact match (EM). For instance, 21.5\% of GPT-3.5's first and second round responses were consistent, whereas this proportion increased to 26.25\% in GPT-4. 
Comparing the first and third round responses, although the similarity score of GPT-3.5 improved, there was still a significant difference in exact match proportions compared to GPT-4, thus validating our hypothesis.
\begin{figure}[htbp]
\centering
\includegraphics[width=0.8\columnwidth]{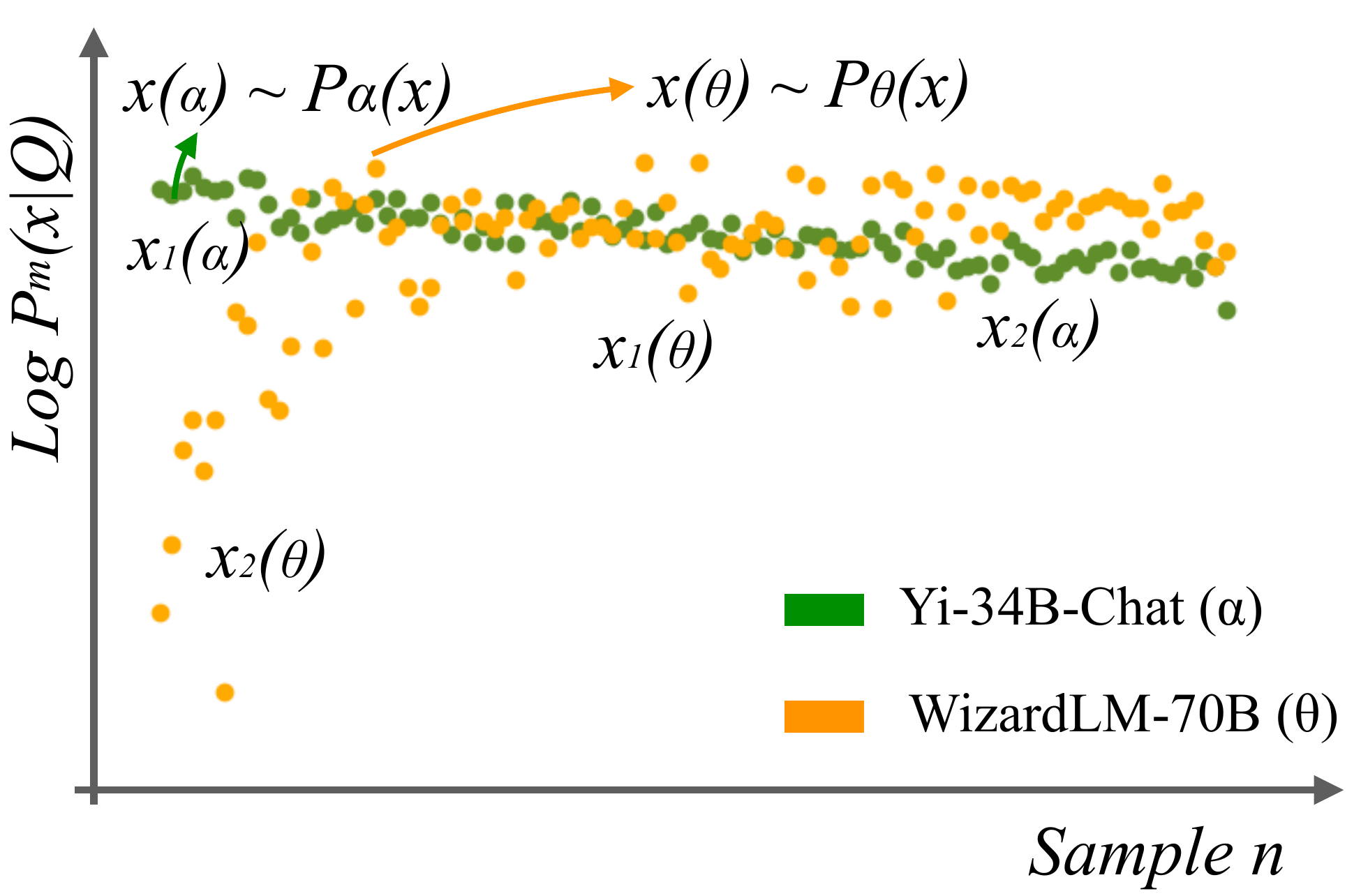}
\caption{Log probability curves of the responses for Yi-34B-Chat and WizardLM-70B on AlpacaEval-2.0.}
\label{fig:log_prob_on_alpacaeval}
\vspace{-0.5em}
\end{figure}

So, why does this phenomenon exist? We posit that it's because stronger LLMs assign higher probabilities to each token in the answer output, resulting in greater determinism and thus smaller variance in the decoded responses. To investigate this conjecture, we conducted a second preliminary study. We selected the AlpacaEval-2.0 benchmark along with the best-performing open-source model on the benchmark, Yi-34B-Chat (Score 29.66\%), and a relatively poorer-performing model, WizardLM 70B (Score 14.38\%). 
We randomly sampled 10 queries from AlpacaEval-2.0, with each query sampled for 100 responses from each model. We plotted the log probability curves of the responses for each model, as shown in Figure \ref{fig:log_prob_on_alpacaeval}. It is evident that the log probability curve of Yi-34B-Chat is flatter, indicating smaller variance, with nearly all response output probabilities being approximately equal.

\begin{figure}[htbp]
\centering
\includegraphics[width=0.8\columnwidth]{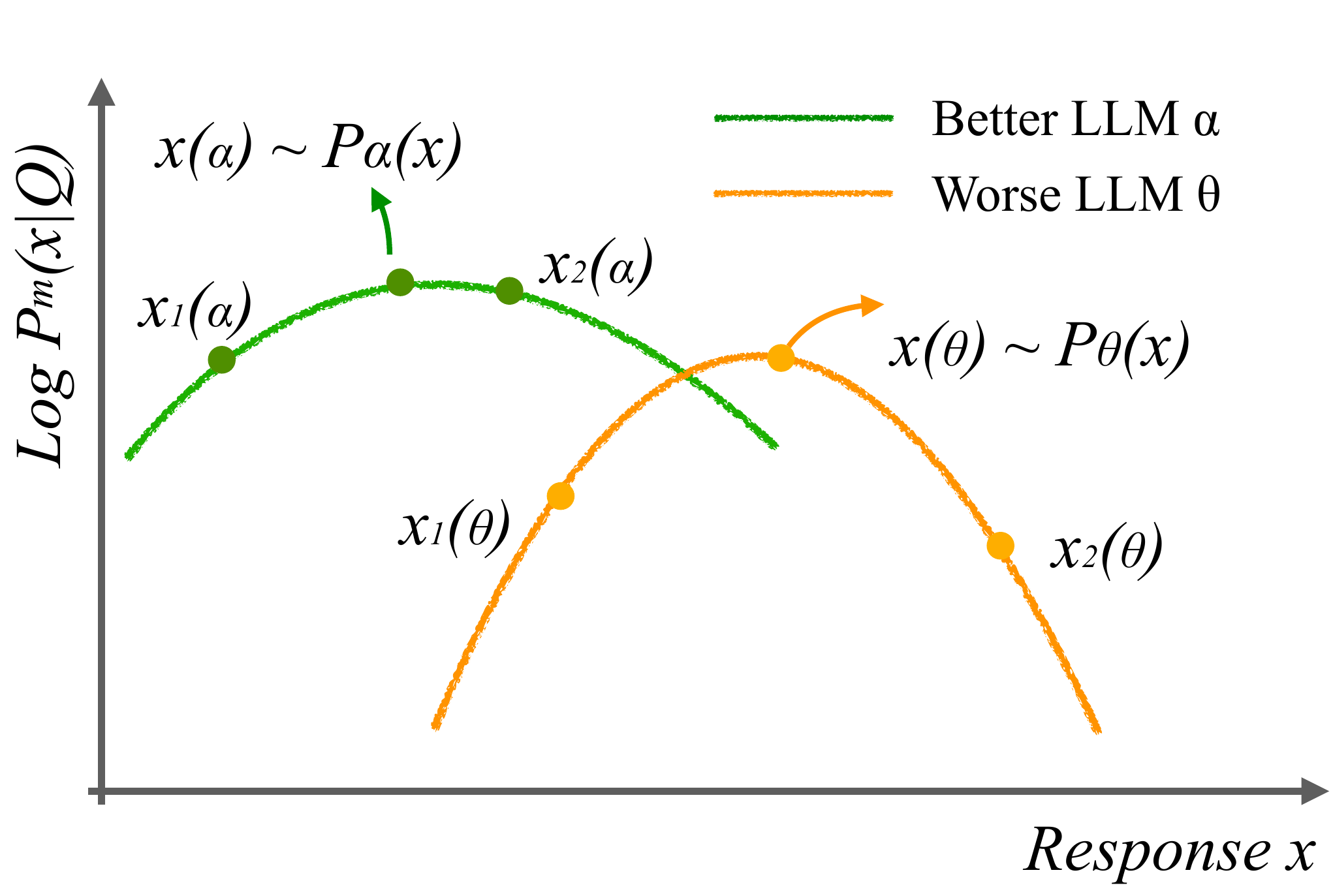}
\caption{We recognize and leverage the observation that superior LLMs typically exhibit smaller probability variances, along with the conclusion that the model-generated samples tend to reside in regions of negative curvature within the probability function. These findings serve as crucial distinctions for ProbDiff in discerning between models of varying capabilities.}
\label{fig:log_curve}
\vspace{-0.5em}
\end{figure}

Combining this conclusion with the observations of DetectGPT~\cite{mitchell2023detectgpt} forms the theoretical foundation of ProbDiff. DetectGPT proposes that samples $x$ from an LLM typically reside in regions of negative curvature of the log probability function, where neighboring samples like $x_1$ and $x_2$ exhibit lower model log probability on average. In other words, the log probability of the responses initially sampled by the model tends to be at local maxima of the probability function.
As illustrated in Figure~\ref{fig:log_curve}, we depict a simulated probability function. 
Due to the fact that the log probability $p(x|q)$ for a specific query $q$ is comparatively flatter for better LLM than for worse LLM, we can infer that models with larger probability discrepancy, after revising their responses multiple times, will perform worse on the current query.

\subsection{Self-evaluation Protocol ProbDiff}

\begin{colblock}[Response Refinement Prompt]

I want you to act as a Response Rewriter.\\
Your goal is to enhance the quality of the response given by an AI assistant to the \#Given Prompt\# through rewriting.\\
But the rewritten response must be reasonable and must be understood by humans.\\
Your rewriting cannot omit the non-text parts such as the emoji in \#Given Prompt\# and \#Given Response\#. \\
If you think the response is already great enough, you can keep it unchanged.\\
You should try your best not to change the length of the response.\\
\#Given Response\# and \#Rewritten Response\# are not allowed to appear in \#Rewritten Response\#.\\
\#Given Prompt\#:\\
\{\textbf{prompt}\}\\
\#Given Response\#:\\
\{\textbf{response}\}\\
\#Rewritten Response\#:
\end{colblock}

Based on our validated assumptions, ProbDiff employs a remarkably simple yet elegant strategy for LLM evaluation. 
Specifically, given any LLM $\alpha$, and a query $q$ under evaluation, we first require $\alpha$ to generate a response $x$ containing $T$ tokens according to its decoding strategy, with the average per-token log probability ${\rm log}; p_{\alpha}(x|q)$ defined as:
\begin{align}
{\rm log}; p_{\alpha}(x|q) = \frac{1}{T}\sum_{t=1}^{T}p_{\alpha}(x_t|q,x_{< t}).
\end{align}
Subsequently, we prompt LLM $\alpha$ to revise the response based on $q$ and $x$, obtaining $x_1$, and iterate this revision process $K$ times to obtain the final response $x_K$, with the revision prompt defined as the Response Refinement Prompt. 
Finally, we calculate the log-probability discrepancy between $x$ and $x_K$ as:
\begin{align}
d(\alpha, q) = {\rm log}\; p_{\alpha}(x_K|q)-{\rm log}\; p_{\alpha}(x|q).
\end{align}
We aim to characterize the variance of LLM $\alpha$'s probability distribution for $q$ using $d(\alpha, q)$, although employing multiple evaluations for averaging could yield more accurate estimates. However, in this paper, we present a preliminary exploration where we adopt the simplest characterization by directly computing the discrepancy between the two probabilities.
Subsequently, for the evaluation set $\mathbb{D}$, we can compute the average $d(\alpha, q)$ for all $q$ in the set, yielding $d(\alpha, \mathbb{D})=\frac{1}{|\mathbb{D}|}\sum_{q\in \mathbb{D}}d(\alpha, q)$ as the probability discrepancy score of LLM $\alpha$ over $\mathbb{D}$. Ultimately, for the two LLMs, $\alpha$ and $\theta$, under evaluation, we can compare $d(\alpha, \mathbb{D})$ and $d(\theta, \mathbb{D})$. A larger $d(\alpha, \mathbb{D})$ implies a higher variance in the probability distribution of the generation, indicating lower confidence in its responses and thus poorer performance under our hypothesis.
Furthermore, if one wishes to statistically analyze the win/tie/lose ratios between $\alpha$ and $\theta$, one can establish predefined thresholds for $d(\alpha, q)$ and $d(\theta, q)$ to determine when their performances are comparable within a certain discrepancy range. Beyond this range, a judgment can be made regarding which model performs better.

In addition, we use Response Refinement Prompt to guide the model to revise the original responses, instead of using multiple sampling to calculate log probability discrepancies. This is mainly because greedy methods are usually used to decode in math and code tasks, such as code and math tasks in MT-Bench. Multiple sampling for this decoding method is meaningless, and using prompt to guide the model to revise the answer is suitable for any decoding methods.



\section{Experiment}

In this section, we evaluate our ProbDiff methodology across three diverse natural language generation tasks: Translation, Summarization, and a niche task designated as Xiaohongshu blog writing. To gauge ProbDiff's efficacy, we fine-tuned the \textit{Qwen-14B-Chat} model for each specified task. Additionally, our evaluation encompasses three benchmarks tailored for LLM assessment: MT-Bench \cite{zheng2023judging}, AlpacaEval \cite{alpaca_eval}, and AlignBench \cite{liu2023alignbench}. These experiments are designed to comprehensively evaluate ProbDiff's performance across a wide array of linguistic tasks and benchmarks, showcasing its applicability in measuring LLM effectiveness.

\begin{table*}[t]
\renewcommand\arraystretch{1} 
\renewcommand\tabcolsep{12.8pt}
\centering
\begin{tabular}{l|ccccccc}
\toprule
\multicolumn{1}{c|}{}
&\textbf{Cs-En}
&\textbf{De-En}
&\textbf{Ru-En} 
&\textbf{Zh-En}
&\textbf{Cnn\_dm}
&\textbf{Xsum}
&\textbf{Blog}\\
\midrule
  


Qwen  &4.6	&1.1	&3.5	&8.1 	&21.5	&17.9 	&64.0 \\

Qwen$_{ft}$ &\textbf{91.3}	&\textbf{94.0}	&\textbf{91.8}	&\textbf{91.4} 	&\textbf{67.1}	&\textbf{77.1} 	&\textbf{90.0} \\

\bottomrule
\end{tabular}
\caption{\textit{Confidence} (\%) obtained through ProbDiff in the translation,  summarization, and Xiaohongshu blog writing tasks. ``Qwen'' represents the Qwen-14B-Chat model with different parameter sizes, and ``Qwen$_{ft}$'' denotes the fine-tuned model based on Qwen-14B-Chat with different parameter sizes. The best performance in each column is bold.} 
\label{tab:nlp_perf}
\end{table*}


\begin{table*}[t]
\renewcommand\arraystretch{1} 
\setlength{\tabcolsep}{9pt}
\centering
\begin{tabular}{l|ccccccccc}
\toprule
{}	&\textbf{Overall} &\textbf{Pro.} &\textbf{Math.} &\textbf{Fund.} &\textbf{Logic.} &\textbf{Writ.} &\textbf{Chi.} &\textbf{Role.} &\textbf{Open.}  \\
	\midrule





Qwen        &32.2	&20.2	&18.9	&15.9	&31.6	&\textbf{62.7}	&13.8	&53.4	&\textbf{42.1} \\
Qwen$_{ft}$ &\textbf{45.7}	&\textbf{25.0}	&\textbf{48.6}	&\textbf{43.5}	&\textbf{62.1}	&52.0	&\textbf{27.6}	&\textbf{60.3}	&34.2 \\

\bottomrule
\end{tabular}
\caption{\textit{Confidence} (\%) obtained through the ProbDiff in AlignBench.  ``Pro.'' denotes
“Professional Knowledge”, ``Math.'' denotes Mathematics, ``Fund.'' denotes Fundamental Language Ability, ``Logic.'' denotes Logical Reasoning, ``Writ.'' denotes Writing Ability, ``Chi.'' denotes Advanced Chinese Understanding, ``Role.'' denotes Task-oriented Role Play, and ``Open.'' denotes
Open-ended Questions. The best performance in each column is bold.} 
\label{tab:llm_perf}
\end{table*}


\subsection{Datasets}
\label{sec:data}

For the \textbf{Translation} task, we selected the \textbf{WMT19} dataset \cite{ng2019facebook} for Chinese-English (\textit{Zh-En}), Czech-English (\textit{Cs-En}), German-English (\textit{De-En}), and Russian-English (\textit{Ru-En}) translation tasks. We randomly chose 5,000 translation pairs for our training sets and 1,000 pairs for our test sets, ensuring an equitable distribution between source-to-English and English-to-source language pairs.
Regarding the \textbf{Summarization} task, we opted for the \textbf{XSum} \cite{xsum-emnlp} and \textbf{CNN\_DM} \cite{hermann2015teaching} datasets, randomly selecting 5,000 documents for the training set and 1,000 documents for the test set from each dataset.


We introduce an innovative \textbf{Xiaohongshu blog writing task} as a novel natural language generation task to assess the instruction-following capabilities of LLMs. For this, we compiled a new benchmark by gathering 569 blogs across various topics from the Xiaohongshu APP, a lifestyle platform where users share product recommendations or opinions. We randomly chose 469 instances for model fine-tuning and 100 instances to evaluate model performance. Detailed descriptions and example instances of the new dataset are available in Appendix \ref{sec:xiaohongshu}.


To evaluate the alignment capability of LLMs across diverse dimensions, we undertook experiments with AlignBench \cite{liu2023alignbench}, a comprehensive dataset designed to assess LLMs' alignment capabilities in Chinese. It categorizes use cases into eight principal domains: Fundamental Language Abilities, Chinese Advanced Understanding, Open-ended Questions, Writing Ability, Logical Reasoning, Mathematics, Task-oriented Role Play, and Professional Knowledge, encompassing 683 samples in total. We used GPT-4 to generate 50 text samples per use case and category pair, and selected 15 generated samples per pair to eliminate duplicates, culminating in a collection of 10,245 generated samples for fine-tuning Qwen-14B-Chat.


In addition to specific task fine-tuning of Qwen-14B-Chat, we also conducted experiments on models listed on the existing LLM benchmark leaderboards.
We choose MT-bench\footnote{\url{https://chat.lmsys.org/?leaderboard}}  and AlpacaEval 2.0\footnote{\url{https://tatsu-lab.github.io/alpaca_eval}} as our evaluation benchmarks. \textbf{MT-bench}\cite{zheng2023judging} is tailored to examine multi-turn conversation and instruction-following capabilities through 80 high-quality multi-turn questions. \textbf{MT-bench}\cite{zheng2023judging} comprises 805 single-turn questions, formulated from Alpaca Data.

\subsection{Implementation Details}



For all experiments except those on the MT-Bench and AlignBench datasets, we configured the temperature to $0.7$, respectively, to generate initial responses. These two benchmarks possess officially recommended generation settings, which we adhered to for relevant experiments. To refine the generated responses, we adjusted the temperature to $0.1$. In translation, summarization, and AlpacaEval tasks, the initial round aimed at general text generation tasks, we set the temperature to $0.7$, referring to the parameters established by MT-Bench. However, for the second round, where we viewed the review of existing answers as a more professional task, we lowered the temperature accordingly.

The effectiveness of an LLM is evaluated based on the confidence level of its responses. This metric quantifies the degree of probability enhancement observed when the model's initial response undergoes refinement. The calculation is as follows:
\begin{align}
confidence =\frac{1}{|\mathbb{D}|}\sum_{q\in \mathbb{D}}\phi
\end{align}
\begin{gather}
    \phi=\left\{
        \begin{array}{lcl}
        1, &d(\alpha, q) \ge \delta\\
        0,  &d(\alpha, q) < \delta
        \end{array} \right. 
        \label{eq:get_explanation}
\end{gather}

where $\delta$ is the threshold. In this paper, $\delta$ is set to $-0.05$. 

\subsection{Results}




\textbf{Natural Language Generation Evaluation}: We applied ProbDiff to assess \textit{Qwen-14B-Chat}'s efficacy in tasks such as translation, summarization, and Xiaohongshu blog writing. Table \ref{tab:nlp_perf} illustrates the \textit{confidence} levels in the generated sentences by both the original and fine-tuned Qwen models across these tasks. 
Subsequent to the fine-tuning process, it was noted that the fine-tuned model exhibited increased confidence in its responses upon revision, aligning with our anticipatory hypotheses.
Specifically, in the Xiaohongshu blog writing task tailored for LLMs with a 14B parameter size, we encountered a scarcity of data for fine-tuning. Nonetheless, post-fine-tuning, there was a noticeable enhancement in response confidence, as depicted in Table \ref{tab:nlp_perf}.
From these results, we deduce that fine-tuning LLMs on task-specific datasets invariably boosts the confidence of their responses, markedly surpassing the performance of the pre-fine-tuned LLM. This elevation in response confidence serves as a pertinent indicator for gauging improvements in LLM performance.



\textbf{LLM's Alignment Evaluation}: To ascertain the alignment capabilities of LLMs across diverse dimensions, we conducted fine-tuning of \textit{Qwen-14B-Chat} using documents synthesized by GPT-4, specifically targeting the AlignBench dataset. The results of our evaluation are presented in Table \ref{tab:llm_perf}.

AlignBench serves as a meticulously curated benchmark designed to assess the alignment capabilities of LLMs. This benchmark poses a more complex challenge compared to standard NLG tasks, requiring nuanced understanding and response accuracy from the model. As depicted in Table \ref{tab:llm_perf}, post fine-tuning with task-specific data, the model demonstrates notable confidence across most evaluated alignment abilities. However, it is observed that in categories such as writing ability and open-ended questions, the model's confidence tends to diminish, suggesting areas for further improvement and investigation.

\begin{table}[h]
\renewcommand\arraystretch{1} 
\setlength{\tabcolsep}{2pt}
\centering
\begin{tabular}{l|cccc}
\toprule
{\multirow{2}{*}{}} 
&\multicolumn{2}{c}{\textbf{MT-Bench}} 
&\multicolumn{2}{c}{\textbf{AlpacaEval}} \\
{} &ProbDiff &Official &ProbDiff &Official \\
\midrule

\textbf{Llama2} &7.50 &6.86 &2.61 &13.87\\
\textbf{Yi} &17.50 &7.58 &6.33 &29.66\\
\textbf{WizardLM} &30.63 &7.71 &4.60 &14.38\\
\textbf{Tulu} &20.00 &7.89 &5.47  &15.98\\

\bottomrule
\end{tabular}
\caption{ProbDiff and Official results on MT-Bench and AlpacaEval. The metrics of ProbDiff is \textit{confidence}(\%), MT-Bench official metrics is model scores(0-10) judged by GPT-4, and AlpacaEval 2.0 official metrics is win rate(\%).} 
\label{tab:mtbench}
\end{table}


\textbf{Evaluation on other LLMs}: To assess the generalization capabilities of our ProbDiff, we extended our evaluations to include several high-performing LLMs featured on established LLM benchmark leaderboards, specifically MT-Bench and AlpacaEval 2.0. We employed ProbDiff to evaluate a selection of models, including Yi-34B-Chat \cite{01yi}, tulu-2-dpo-70B \cite{ivison2023camels}, WizardLM-70B-V1.0 \cite{xu2023wizardlm}, and Llama2-70B-chat \cite{touvron2023llama}. Table \ref{tab:mtbench} presents the response confidence of these LLMs alongside their official benchmarks. Notably, Yi-34B-Chat lacks official MT-Bench scores, thus we conducted its evaluation using the benchmark's official protocol.

\begin{figure}[htbp]
\small
\centering
\includegraphics[width=0.9\columnwidth]{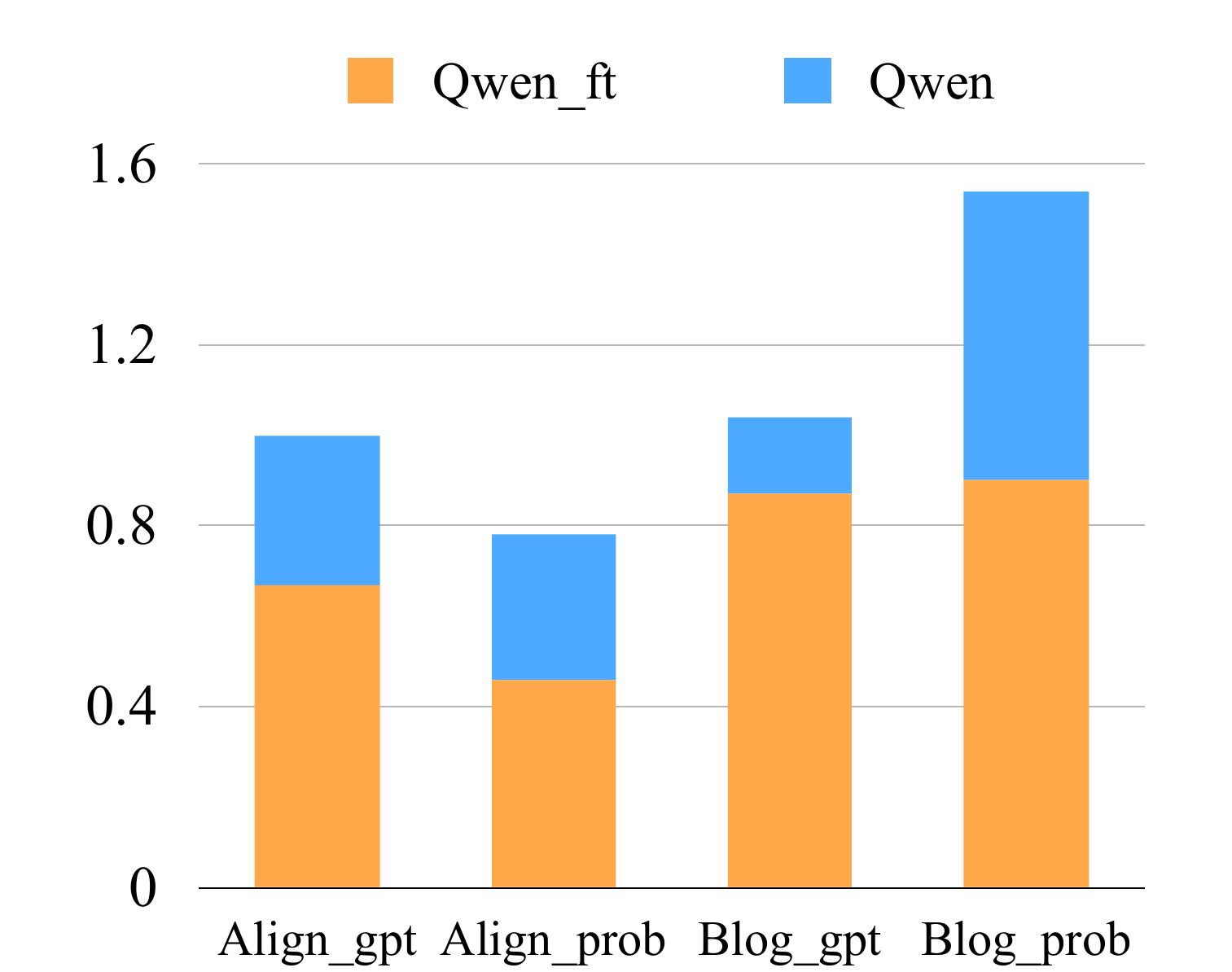}
\caption{Evaluate the validity of the Qwen-14B-Chat and Qwen-14B-Chat\_{ft} through GPT-4 in AlignBench(Align) and Xiaohongshu Blog Writing(Blog) tasks. ``Align\_gpt'' and ``Blog\_gpt'' represents the win rate judged by GPT-4, ``Align\_prob'' and ``Blog\_prob'' represents the confidence evaluate by ProbDiff. Orange histogram indicates fine-tuned Qwen-14B-Chat and blue histogram indicates Qwen-14B-Chat, respectively.}
\label{fig:gpt4-a}
\vspace{-0.35em}
\end{figure}

\begin{table*}[t]
\renewcommand\arraystretch{1.3} 
\renewcommand\tabcolsep{5pt}
\centering\small
\begin{tabular}{ll|cccccccccccc}
\toprule
\multicolumn{2}{c|}{\multirow{2}{*}{}} 
&\textbf{Overall} &\textbf{Pro.} &\textbf{Math.} &\textbf{Fund.} &\textbf{Logic.} &\textbf{Writ.} &\textbf{Chi.} &\textbf{Role.} &\textbf{Open.}  \\
\midrule
    
\multirow{2}{*}{\textbf{GPT-4}} 
&Qwen &5.66	&\underline{6.28}	&4.89	&7.00	&4.66	&6.56	&\underline{6.17}	&6.47	&6.74 \\
&Qwen$_{ft}$ &\underline{6.60}	&6.27	&\underline{5.85}	&\underline{7.15}	&\underline{7.12}	&\underline{7.00}	&5.83	&\underline{7.16}	&\underline{6.89} \\

\multirow{2}{*}{\textbf{CritiqueLLM}} 
&Qwen         &5.22	&\underline{5.90}	&4.23	&\underline{6.43}	&4.28	&6.48	&\underline{5.97}	&6.03	&6.34 \\
&Qwen$_{ft}$  &\underline{6.19}	&5.58	&\underline{5.42}	&6.28	&\underline{6.98}	&\underline{6.77}	&4.97	&\underline{6.99}	&\underline{6.45} \\

\multirow{2}{*}{\textbf{ProbDiff}} 
 &Qwen  &32.2	&20.2	&18.9	&15.9	&31.6	&\underline{62.7}	&13.8	&53.4	&\underline{42.1} \\
\multirow{2}{*}{}   &Qwen$_{ft}$ &\underline{\textbf{45.7}}	&\underline{25.0}	&\underline{\textbf{48.6}}	&\underline{\textbf{43.5}}	&\underline{\textbf{62.1}}	&52.0	&\underline{27.6}	&\underline{\textbf{60.3}}	&34.2 \\

\bottomrule
\end{tabular}
\caption{The scores(0-10) obtained through GPT-4 and CritiqueLLM of AlignBench, and the response confidence(\%) obtained through ProbDiff. ``Qwen'' represents the Qwen-14B-Chat model, and ``Qwen$_{ft}$'' denotes the fine-tuned model based on Qwen-14B-Chat. The underlined number indicates the best performance evaluated by different approaches. The bold numbers indicate the ProbDiff results consistent with GPT-4.} 
\label{tab:align}
\end{table*}

Within the MT-Bench framework, the official ranking of these LLMs is Tulu, WizardLM, Yi, and Llama2, a sequence that aligns perfectly with the model confidence hierarchy as determined by ProbDiff, as illustrated in table \ref{tab:mtbench}. Conversely, for MT-Bench, the ranking is Tulu, WizardLM, Yi, and Llama2. Our analysis, however, positions the models as WizardLM, Tulu, Yi, and Llama2, showcasing a slight variation from the official results. 
This discrepancy suggests that ProbDiff’s criteria for evaluating multi-round dialog tasks may still harbor certain imperfections.

\section{Discussion}
\subsection{Confidence vs GPT-4 scores}


In this section, to substantiate our premise that enhancements in model efficacy are discernible through shifts in log probability, we juxtaposed the congruence between our evaluative outcomes and those derived from utilizing GPT-4 as an adjudicator. To juxtapose the evaluative outcomes of ProbDiff with those of GPT-4, we engaged GPT-4 to directly appraise which output from the two models prevails, eschewing the assignment of a definitive score. Figure \ref{fig:gpt4-a} illustrates that when ProbDiff adjudicates the fine-tuned model as superior to its baseline counterpart, GPT-4's evaluation of the outputs from these models aligns with the determinations made by ProbDiff.


Given that AlignBench furnishes official evaluation scripts capable of producing scores for each model via GPT-4 or alternative evaluative LLMs like CritiqueLLM \cite{ke2023critiquellm}, we have also juxtaposed these findings in Table \ref{tab:align}. To undertake a comprehensive assessment, AlignBench articulates a detailed taxonomy of LLM capabilities, categorizing and summarizing their applications across 8 principal categories. The specifics of this comparative analysis are presented in Table \ref{tab:align}. It's observable that while disparities between the outcomes derived from ProbDiff and those obtained via GPT-4 manifest in certain evaluative contexts, such as writing ability, our ProbDiff methodology consistently aligns with the evaluations conducted by GPT-4 and CritiqueLLM in the majority of instances.

\subsection{ProbDiff vs GPT-4 and Manual Evaluation}


In soliciting amendments from LLMs, we employ the log probability variance as a gauge to ascertain the enhancement of the refined responses. This method involves a rigorous examination of the congruence between ProbDiff, GPT-4, and manual evaluations by quantifying the extent of discordance. It is essential to articulate that, from a probabilistic vantage, the model invariably selects the response with the maximal probability. Upon requesting an LLM to adjust its outputs, should minute modifications materialize, such as the substitution of synonyms, there ensues a marked diminution in the log probability of the adjusted responses. Nevertheless, considering the dimension of response quality, these minimal alterations do not compromise the aggregate quality of the response. Consequently, both GPT-4 and manual evaluations do not perceive these minor adjustments as significant enough to impact the overall quality of the response. Hence, probability emerges as a stringent metric, culminating in a diminished correlation coefficient among ProbDiff, GPT-4, and manual evaluations.

\begin{table}[h]
\renewcommand\arraystretch{1} 
\setlength{\tabcolsep}{8pt}
\small
\centering
\begin{tabular}{l|cc}
\toprule
{\multirow{2}{*}{}} 
&\textbf{GPT-4}
&\textbf{Humans} \\
\midrule

\textbf{AlignBench} &3.84 &4.49\\
\textbf{MT-Bench} &6.45 &4.84\\

\bottomrule
\end{tabular}
\caption{Conflict degree(\%) among ProbDiff, GPT-4, and manual evaluation. We employ 3 human annotators to report the average results.} 
\label{tab:conflict}
\end{table}

We deployed GPT-4 alongside manual evaluation to ascertain whether our ProbDiff engenders outputs that diverge from the evaluations provided by GPT-4 and manual review. Our objective was to scrutinize instances where ProbDiff indicates an improvement ($d(\alpha, q)>0$), to determine if both GPT-4 and human assessors concurrently affirm an enhancement in response quality. Conversely, in scenarios where ProbDiff suggests a decline ($d(\alpha, q)<0$), we aimed to validate whether GPT-4 and human evaluators corroborate a diminution in response quality. This investigation was carried out on AlignBench and MT-Bench, with outcomes detailed in table \ref{tab:conflict}. As elucidated in table \ref{tab:conflict}, the discrepancy between ProbDiff and the combined assessments of GPT-4 and manual evaluation remains minimal. Our human evaluation was conducted by 3 Ph.D. volunteers specializing in NLP. These volunteers assessed whether the quality of the modified answers improved. If the quality remained unchanged, 0 points will be assigned. If the quality improved 1 point was awarded, and if it decreased -1 point was given. The average kappa coefficient among the human annotators was 0.66 on AlignBench and 0.62 on MTBench.

\section{Conclusion}

In this paper, we introduce an innovative evaluation framework, ProbDiff, designed to facilitate LLMs in performing self-assessment through the analysis of probability discrepancies in their generated outputs. Furthermore, our ProbDiff method offers a robust mechanism for evaluating the competencies of LLMs, even without pre-existing training or domain-specific benchmarks, effectively addressing concerns of data confidentiality breaches.

\section*{Acknowledgement}
The authors would like to thank the anonymous referees for their valuable comments. 
This work is supported by the National Science and Technology Major Project under Grant No.2023YFF0905400, 
the National Natural Science Foundation of
China (No.U2341229),  and the Fundamental Research Funds for the Central Universities, JLU.

\section*{Limitations}

The constraints of our methodology are outlined as follows:
Firstly, while our approach is capable of ascertaining enhancements in LLM performance, it lacks the capacity to offer quantitative metrics indicating the magnitude of improvement. ProbDiff is more focused on giving log-probability based trends, independent of external models, rather than yielding precise results. Consequently, ProbDiff is more suitable as an auxiliary evaluation method to help researchers view the changing trends of model effects. Secondly, variations in log probability are occasionally affected by sentence length. Should there be a notable disparity in length between the original and modified sentences, the log probability will undergo substantial changes.

\bibliography{custom}
\bibliographystyle{acl_natbib}

\newpage
\appendix

\section{Xiaohongshu Datasets}
\label{sec:xiaohongshu}
\begin{figure}[htbp]
\centering
\includegraphics[width=\columnwidth]{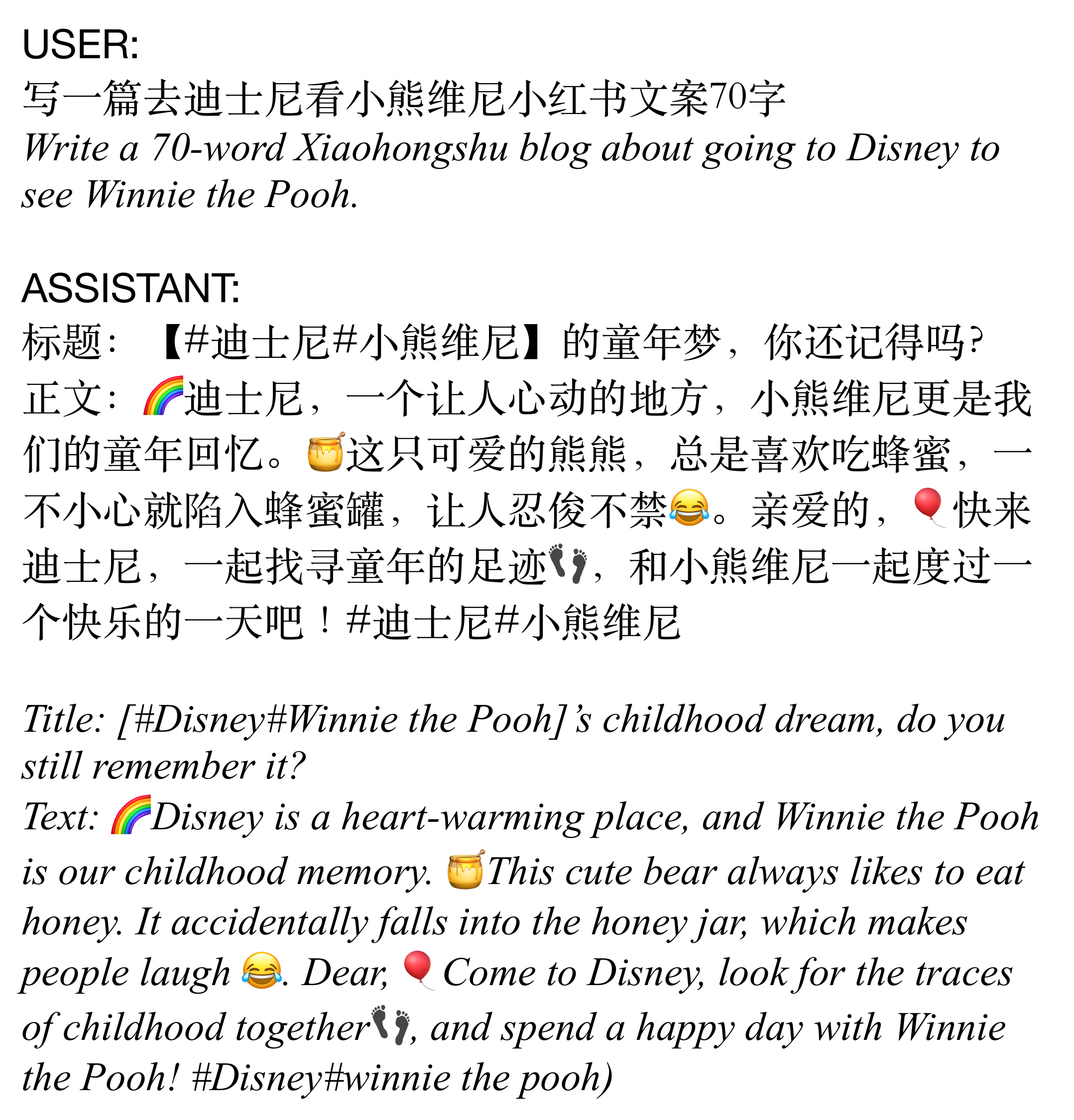}
\caption{Xiaohongshu blog writing data.}
\label{fig:redbook}
\vspace{-0.5em}
\end{figure}

According to the blog format and text style of Xiaohongshu App, we manually wrote 569 blogs, covering various functional topics such as life, beauty, travel, recommendations, etc., and constructed a instruction following dataset. It is characterized by the addition of rich emoji elements. The data format is shown in figure \ref{fig:redbook}

\section{Threshold analysis}
In this section, we analyze the impact of different thresholds on response confidence. We conducted experiments in Alignbench and set the thresholds to 0, -0.05, and -0.1 to observe the impact of the thresholds on the experimental results.

As shown in table \ref{tab:llm_perf1}, when the absolute value of the threshold decreases, although the final results of most sub-tasks are consistent with those of GPT-4, for untuned models, lower absolute values can significantly affect the confidence in certain sub-tasks, such as mathematics, fundamental language, and role-play. As the absolute value of the threshold increases, the model's performance deviates more from the GPT-4 evaluation results. Therefore, we chose -0.05 as the final threshold.

\begin{table*}[t]
\renewcommand\arraystretch{1.35} 
\setlength{\tabcolsep}{7.25pt}
\centering
\begin{tabular}{ll|ccccccccc}
\toprule
	 \multicolumn{2}{c|}{\multirow{2}{*}{}} 
	&\textbf{Overall} &\textbf{Pro.} &\textbf{Math.} &\textbf{Fund.} &\textbf{Logic.} &\textbf{Writ.} &\textbf{Chi.} &\textbf{Role.} &\textbf{Open.}  \\
	\midrule

\multirow{2}{*}{\textbf{0}} &Qwen &18.7	&10.5	&6.3	&8.7	&23.2	&37.3	&6.9	&37.1	&13.2 \\
\multirow{2}{*}{}  &Qwen$_{ft}$ 
&\underline{41.6}	&\underline{25.0}	&\underline{43.2}	&\underline{40.6}	&\underline{55.8}	&\underline{41.3}	&\underline{27.6}	&\underline{56.0}	&\underline{31.6} \\

 \multirow{2}{*}{\textbf{-0.05}} &Qwen  &32.2	&20.2	&18.9	&15.9	&31.6	&\underline{62.7}	&13.8	&53.4	&\underline{42.1} \\

 \multirow{2}{*}{}  &Qwen$_{ft}$ &\underline{45.7}	&\underline{25.0}	&\underline{48.6}	&\underline{43.5}	&\underline{62.1}	&52.0	&\underline{27.6}	&\underline{60.3}	&34.2 \\

\multirow{2}{*}{\textbf{-0.1}} &Qwen &41.1	&26.6	&29.7	&26.1	&34.7	&\underline{72.0}	&20.7	&\underline{65.5}	&\underline{57.9} \\
\multirow{2}{*}{}  &Qwen$_{ft}$ &\underline{50.5}	&\underline{27.4}	&\underline{55.0}	&\underline{44.9}	&\underline{70.5}	&58.7	&\underline{29.3}	&62.9	&47.4 \\

\bottomrule
\end{tabular}
\caption{\textit{Confidence} (\%) obtained through the ProbDiff in AlignBench with different $\delta$.  ``Pro.'' denotes
“Professional Knowledge”, ``Math.'' denotes Mathematics, ``Fund.'' denotes Fundamental Language Ability, ``Logic.'' denotes Logical Reasoning, ``Writ.'' denotes Writing Ability, ``Chi.'' denotes Advanced Chinese Understanding, ``Role.'' denotes Task-oriented Role Play, and ``Open.'' denotes
Open-ended Questions. } 
\label{tab:llm_perf1}
\end{table*}


\begin{table*}[t]
\renewcommand\arraystretch{1} 
\setlength{\tabcolsep}{9pt}
\centering
\begin{tabular}{l|ccccccccc}
\toprule
{}	&\textbf{Overall} &\textbf{Pro.} &\textbf{Math.} &\textbf{Fund.} &\textbf{Logic.} &\textbf{Writ.} &\textbf{Chi.} &\textbf{Role.} &\textbf{Open.}  \\
	\midrule

Qwen         &43.8	&26.6	&41.4	&29.0	&44.2	&\textbf{66.7}	&12.1	&\textbf{64.7}	&\textbf{68.4} \\
Qwen$_{ft}$  &\textbf{52.7}	&\textbf{38.7}	&\textbf{60.4}	&\textbf{59.4}	&\textbf{58.9}	&57.3	&\textbf{34.5}	&56.9	&50.0 \\

\bottomrule
\end{tabular}
\caption{\textit{Confidence} (\%) obtained through the ProbDiff with the simple prompt in AlignBench.   The best performance in each column is bold.} 
\label{tab:prompt_analysis}
\end{table*}


\begin{table*}[!hbp]
\renewcommand\arraystretch{1} 
\setlength{\tabcolsep}{7.25pt}
\centering
\begin{tabular}{ll|ccccccccc}
\toprule
\multicolumn{2}{c|}{\multirow{2}{*}{}} 
&\textbf{Overall} &\textbf{Pro.} &\textbf{Math.} &\textbf{Fund.} &\textbf{Logic.} &\textbf{Writ.} &\textbf{Chi.} &\textbf{Role.} &\textbf{Open.}  \\

\midrule

\multirow{2}{*}{\textbf{same}} &Qwen &23.0	&20.2	&18.9	&15.9	&31.6	&38.7	&10.3	&25.9	&13.2 \\
\multirow{2}{*}{}  &Qwen$_{ft}$ 
&\textbf{41.6}	&\textbf{25.0}	&\textbf{48.6}	&\textbf{43.5}	&\textbf{62.1}	&\textbf{44.0}	&\textbf{27.6}	&\textbf{43.1}	&\textbf{28.9} \\

 \multirow{2}{*}{\textbf{0.7}} &Qwen  &12.7	&3.2	&7.2	&7.2	&12.6	&26.7	&1.7	&28.4	&10.5\\

 \multirow{2}{*}{}  &Qwen$_{ft}$ &\textbf{37.5}	&\textbf{17.7}	&\textbf{34.2}	&\textbf{31.9}	&\textbf{57.9}	&\textbf{46.7}	&\textbf{22.4}	&\textbf{51.7}	&\textbf{28.9}\\

\bottomrule
\end{tabular}
\caption{\textit{Confidence} (\%) obtained through the ProbDiff with the simple prompt in AlignBench.   The best performance in each column is bold. `` same'' indicates the second round temperatures consistent with the first round, 0.7 indicates the second round temperature is set to 0.7.} 
\label{tab:temp_analysis}
\end{table*}

\section{Refinement Prompt analysis}
In this section, we analyze the impact of different prompts on response confidence. We conducted experiments in Alignbench and choose a simple prompt  to observe the impact of the prompts on the experimental results.

\begin{colblock}[Simple Refinement Prompt]
According to the given \#INSTRUCTION\#, please modify the given \#RESPONSE\# to make it better.\\
\#INSTRUCTION\#:\\
\{instruction\}\\
\#RESPONSE\#:\\
\{response\}\\
\#Your Response\#:
\end{colblock}

As shown in table \ref{tab:prompt_analysis}, it is evident that while the confidence levels of the models increase with the implementation of the simple refinement prompt, the overarching trends remain largely consistent with the initial findings, with the exception of role play as delineated in table \ref{tab:llm_perf}.

\section{Refinement Temperature analysis}
In this section, we analyze the impact of different temperature on response confidence. As for Alignbench, the official recommend inference temperature is 0.7 for writing, role play and open-ended questions and 0.1 for other sub-tasks. We kept the 2nd round temperature consistent with the first round in different sub-tasks on AlignBench, and set the 2nd round temperature to 0.7. The results are shown in table  \ref{tab:temp_analysis}. Overall, although there will be some differences in a small number of tasks, changes in temperature will not significantly affect the evaluation results. Additionally, when higher temperature is employed to alter the model's initial round of responses, fine-tuned models tend to exhibit heightened confidence, even in areas where the model may not have received robust training. Therefore, we recommend employing lower temperature when modifying responses, as it yields more stable results during evaluating.

\end{document}